\documentclass[10pt]{article}
\usepackage[letterpaper,margin=1in]{geometry}
\usepackage[T1]{fontenc}
\usepackage{lmodern}
\usepackage{microtype}
\usepackage{amsmath,amsfonts}
\usepackage[authoryear,round]{natbib}
\usepackage{graphicx}
\usepackage{booktabs}
\usepackage{tabularx}
\usepackage{array}
\usepackage{url}
\usepackage{hyperref}
\hypersetup{
  hidelinks,
  pdftitle={Decision Shifts, Lost Label Functionality, and an Inconclusive Grounding Audit in Correctness-Gated Multi-Teacher Distillation},
  pdfauthor={Xiaofei Feng}
}
\title{Decision Shifts, Lost Label Functionality,\\ and an Inconclusive Grounding Audit\\ in Correctness-Gated Multi-Teacher Distillation\thanks{AI-use disclosure: OpenAI Codex was used on 30 August and 9 September 2026 to assist with manuscript restructuring, consistency checking, language editing, and format conversion. It was not assigned authorship. The author remains responsible for verifying the scientific claims, citations, numerical results, and final submitted text.}}
\author{Xiaofei Feng\\
University of Illinois Urbana-Champaign\\
\texttt{xfeng18@illinois.edu}}
\date{}
\begin{document}
\maketitle
\begin{abstract}
Candidate decision correctness and rationale grounding are different objectives. We examine how a correctness gate changes selective decisions and whether the available evidence supports a grounding claim in a fixed multi-teacher distillation experiment. All eight arms share 4,330 sources, a 63.9M-parameter student, 12,990 optimization rows, 406 updates, evidence inputs, and a decoder; the seven teacher-based arms draw from one fixed three-response pool. Three seeds are evaluated on 267 held-out examples. Relative to unfiltered multi-teacher distillation, the correctness-gated weighted arm differed on three fixed-operating-point decision summaries: accuracy \(+0.1660\), 95\% observed-matrix interval \([0.0670,0.2455]\), five-label macro-F1 \(+0.1323\), \([0.0916,0.1731]\), and task-defined conditional unsafe-action rate \(-0.4979\), \([-0.5926,-0.3686]\). These shifts reflect a different decision policy, not uniformly better task behavior. Source-label SFT had the highest mean five-label macro-F1 (0.586). The weighted arm had zero Refuted recall in every seed, and two seeds assigned NotEnoughInfo to all 167 claim examples. In an availability-amended audit at one reference seed, weighted and unfiltered outputs respectively had 0/20 versus 1/20 evidence-supported positives and 20/20 versus 19/20 positives containing unsupported material. The samples were non-paired, source overlap was not serialized, and the amendment occurred after the automatic summary but before annotation; the audit therefore cannot estimate a common-source grounding effect and is inconclusive about system-level improvement or harm. The hard-filter arm already achieved 0.660 accuracy, 0.530 macro-F1, and 0.135 conditional unsafe rate. The implemented weighted arm showed no demonstrated incremental decision benefit over hard filtering. This fixed-matrix failure analysis shows a decision redistribution with lost label functionality, while the available human audit does not establish a grounding gain.
\end{abstract}

\section{Introduction}
Knowledge distillation compresses behavior from a larger teacher into a smaller student \citep{hinton2015distilling}. Rationale distillation adds natural-language intermediate supervision \citep{shridhar2023reasoning,wang2023scott,lee2024mentorkd}, and such supervision can improve data efficiency for compact models \citep{hsieh2023distilling}. Multi-teacher variants provide candidate rationale traces or supervision from multiple pretrained language-model teachers \citep{tian2025beyond,wu2021oneteacher}. This additional supervision is attractive when latency, privacy, or hardware constraints preclude a large model at inference time. Its usefulness, however, depends on what the student is asked to imitate.

Evidence-grounded decisions expose a distinction that ordinary answer accuracy can hide. A claim should be labeled Supported or Refuted only when the supplied evidence licenses that polarity. Otherwise the appropriate decision is NotEnoughInfo. A recommendation query should be answered only when the supplied evidence is adequate. Otherwise the appropriate decision is Abstain. The same output may contain a correct decision and an unsupported rationale. Conversely, a system may reduce false positive decisions by assigning a conservative label to almost every example. Neither behavior establishes grounded reasoning by itself.

Quality-aware distillation methods select teachers, reject candidates, or vary supervision weights \citep{yuan2021reinforced,ding2024tradeoff,li2025committee}. Rationale-specific methods score quality, uncertainty, or student need \citep{wang2025qcrd,zhang2024elad,yan2025efficient,song2025rationale}. Verifiers can rank solutions and support correction \citep{hosseini2024vstar,kawabata2024rationaleaware,zhang2024strongverifiers,dixit2026aletheia}. These results show that teacher outputs need not receive equal trust. They do not imply that a gate trained for candidate decision correctness will improve the evidential support of the language generated by a student.

We study that gap in one controlled experiment. For each training source, three fixed teacher responses form a candidate family. A composite gate estimates whether a candidate's decision matches a human training label. The primary intervention removes candidates below a threshold and weights the retained candidates by the estimated score. Seven comparison arms separate direct source-label training, final-answer transfer, single-teacher rationale transfer, unfiltered multi-teacher transfer, hard filtering, an alternative product weight, and a role-aware gate. All arms use a common student architecture, evidence inputs, row count, update count, and decoder.

The primary comparison is correctness-weighted multi-teacher distillation (MTD) minus uniform unfiltered MTD. We ask two questions. First, do the arms differ on fixed decision endpoints: accuracy, five-label macro-F1, and a narrowly defined conditional unsafe-action rate? Second, do shifts toward higher aggregate scores coexist with functional recall for every label and evidence-supported positive outputs? A grounding interpretation requires both. Automatic support diagnostics cannot substitute for human judgments, and a lower conditional unsafe-action rate cannot compensate for zero or near-zero correct recovery of Supported or Refuted examples.

The study makes three contributions. First, it reports a controlled fixed-matrix failure analysis in which aggregate decision endpoints shift while label functionality is lost and the available grounding audit remains inconclusive. Second, the retained-set- and decision-matched hard arm already exhibits the aggregate pattern, while the implemented weighted arm shows no demonstrated incremental decision benefit; target-text differences preclude a scalar-only or mediator claim. Third, the evaluation combines the highest-macro-F1 source-label control, per-label recall at each observed seed, narrow unsafe-action accounting, and a system-label-masked, availability-amended audit. These checks reveal behavior that aggregate accuracy alone would conceal.

The conclusion is deliberately bounded. We do not propose correctness-weighted distillation as a generally superior method, and we do not conclude that verifier-guided selection is ineffective in other settings. The evidence concerns one student, one preserved teacher pool, one English restaurant-evidence domain, one held-out set, and one decision policy.

\section{Related Work}
\subsection{Rationale and multi-teacher distillation}
Rationale distillation supervises compact models with intermediate explanations in addition to task labels \citep{hsieh2023distilling,shridhar2023reasoning,wang2023scott,lee2024mentorkd}. Recommendation-specific work likewise transfers rationale-bearing outputs to smaller models \citep{wang2024rdrec}. Multi-teacher approaches expose a student to candidate rationale traces or multiple pretrained language-model teachers \citep{tian2025beyond,wu2021oneteacher}, but teacher quantity or capacity does not improve student performance monotonically \citep{ding2024tradeoff}. This distinction is important here: \citet{wu2021oneteacher} supports distillation from multiple pretrained teachers, whereas the non-monotonic teacher-ensemble claim is supported by \citet{ding2024tradeoff}.

Multi-teacher methods intervene at different points. Reinforced selection learns an instance-dependent teacher policy \citep{yuan2021reinforced}. Committee-based methods use peer review and thresholding to reject candidates \citep{li2025committee}. Other work modifies rationale contrast, selection, or weighting \citep{wang2025qcrd,zhang2024elad,yan2025efficient,song2025rationale}. Selective knowledge distillation also filters sequence-level targets \citep{liu2023selective}. These approaches motivate quality-aware supervision, but candidate removal and continuous weighting are not equivalent operations. Our hard and weighted arms retain the same candidate IDs, providing a retained-set- and decision-matched comparison; because some soft targets and serialized conversations differ, the contrast does not isolate continuous mass allocation.

\subsection{Verifiers and objective alignment}
Verifiers can rank candidate solutions and support correction \citep{hosseini2024vstar,zhang2024strongverifiers}. Verified repair traces can also supervise a compact model \citep{dixit2026aletheia}. Rationale-aware verification shows that answer correctness and reasoning validity can diverge \citep{kawabata2024rationaleaware}, while natural-language critics can improve a verifier by identifying reasoning errors \citep{gao2025critics}. The target used by our gate is narrower: a structurally valid candidate receives a positive target when its decision equals the adjudicated training label. Verifier compatibility is only one feature family among agreement, confidence, structural validity, task, decision, and model identity. The intervention is therefore a composite candidate-correctness gate, not a clean causal test of verifier features and not a direct grounding supervisor.

This target distinction creates an objective-alignment problem. A candidate may predict Supported correctly while adding an uncited or overgeneralized material claim. A query may select Answer correctly while its rationale misstates the evidence. Optimizing candidate decision correctness may improve decision targets without improving every language-model target that accompanies them. The present study tests that possibility downstream rather than assuming that a gate score transfers its intended semantics to the student.

The closest work spans complementary axes rather than the same evidential comparison. Multi-teacher and rationale-transfer studies motivate candidate traces and multiple teachers \citep{tian2025beyond,wu2021oneteacher}; selection and committee methods motivate rejecting or reweighting candidates \citep{yuan2021reinforced,li2025committee}; rationale-aware verification separates answer correctness from reasoning validity \citep{kawabata2024rationaleaware}. Our incremental contribution is narrower: within one frozen matrix, we combine a decision-correctness target, a retained-candidate-set-matched hard-versus-weighted comparison, per-label operating-point checks, and a system-label-masked human audit. We claim this combination as a controlled failure analysis, not priority over those methods or a new general-purpose algorithm.

\subsection{Evidence sufficiency and selective prediction}
Evidence insufficiency should remain distinct from either factual polarity \citep{atanasova2022insufficient}. Selective question answering and abstention research similarly evaluate when a model should decline to answer \citep{xin2021abstention,feng2024abstain,madhusudhan2025notanswer,kamath2020selective}. Selective classification formalizes error-coverage trade-offs and asymmetric rejection \citep{geifman2017selective,gangrade2021selective}. Context-adaptive abstention policies further emphasize that a reported risk value belongs to a specified operating rule \citep{tayebati2025cap}.

Probability fit and operational selectivity are also separate. Post-hoc temperature scaling can reduce confidence miscalibration \citep{guo2017calibration}; it does not by itself choose a task-appropriate operating point or guarantee label coverage at a fixed threshold. We therefore report five-label macro-F1 and per-label recall alongside the task-defined conditional unsafe-action rate. Human evidence review is restricted to emitted positive outputs because that is where answer support, rationale faithfulness, unsupported material, and citation validity are applicable. This joint view prevents a conservative decision policy from being mistaken for evidence-grounded competence.

\section{Task, Data, and Teacher Materials}
\subsection{Evidence-grounded tasks and labels}
Each example contains a task type, a claim or recommendation query, a fixed set of temporally indexed evidence snippets, and a leakage-group identifier. Claim verification has three mutually exclusive legal labels:

\[
\mathcal{Y}_{c}=\{\mathrm{Supported},\mathrm{Refuted},\mathrm{NotEnoughInfo}\}.
\]

Recommendation decisions have two:

\[
\mathcal{Y}_{q}=\{\mathrm{Answer},\mathrm{Abstain}\}.
\]

A positive output includes a decision, a natural-language answer or verdict, a rationale, and cited evidence identifiers. Retrieval results, evidence text, temporal snapshots, and split assignments are held fixed across the formal arms. The experiment thus evaluates supervision construction and student behavior over supplied evidence, not retrieval.

The task-defined pooled conditional unsafe-action rate has a narrow numerator and denominator. For claims, an unsafe action is Supported or Refuted only when the gold label is NotEnoughInfo. The task-local denominator is the gold-NotEnoughInfo subset. For queries, an unsafe action is Answer when the gold action is Abstain. The denominator is the gold-Abstain subset. The pooled summary aggregates these safe-required cases. Supported-for-Refuted and Refuted-for-Supported are ordinary classification errors and do not enter this unsafe numerator. The metric is not a general measure of harmful language or deployment safety.

\subsection{Human-gold corpus and split boundaries}
The active human-adjudicated corpus contains 882 source examples: 353 train, 262 legacy development, and 267 held-out test examples. The training split contains 253 claims and 100 queries, with label counts 83 Supported, 53 Refuted, 117 NotEnoughInfo, 34 Answer, and 66 Abstain. Legacy development contains 40, 30, 92, 31, and 69 examples in the same label order. The test contains 167 claims and 100 queries: 44 Supported, 34 Refuted, 89 NotEnoughInfo, 33 Answer, and 67 Abstain.

The restaurant evidence derives from the Yelp Open Dataset, whose official description identifies reviews and business metadata and states an educational-use intent \citep{yelp2026opendataset}. The archived study record does not preserve the acquired release identifier, effective governing terms, or a publication/release determination. This source citation establishes provenance only; it does not establish permission to redistribute derived text or labels.

Legacy development had been exposed during early exploratory work and was not used for formal student, prompt, threshold, or checkpoint selection. Held-out results use 22 leakage groups as the cluster unit for resampling. A separate split audit found no train-development-test intersections for the recorded entity, review, evidence, or normalized text signatures. That audit does not remove the within-training gate-fold limitation described in Section 4.1.

\subsection{Student-training pool and teacher responses}
All arms use a common pool of 4,330 student-training sources: 353 human-gold training sources and 3,977 weak or silver sources. The task composition is 3,598 claim-verification and 732 query-decision sources. Source-label SFT therefore uses the available source labels, not 4,330 independently human-adjudicated labels. The 353 human sources are a subset of this pool.

The archived record identifies the 3,977 non-gold sources as weak or silver supervision but does not provide a complete source-level account of how each reference label was generated, calibrated, or quality-controlled. We therefore treat this mixture as a fixed experimental input rather than as equivalent to human annotation. Label noise or provenance heterogeneity in this majority component may influence every arm, including the source-label SFT control.

Each source has three fixed diagonal teacher responses, giving 12,990 candidates. The recorded Ollama mapping is recommendation to command-r7b:latest, verification to qwen3.5:9b-q4\_K\_M, and abstention to granite4.1:8b-q5\_K\_M \citep{ollama2026command,ollama2026qwen,ollama2026granite}. Appendix B gives the fixed prompts and generation settings.

Teacher generation preceded later adjudication. A later record changed label or provenance metadata for 750 request rows while preserving all 12,990 teacher-visible messages; teachers were not regenerated. The model names are mutable tags, and blob hashes, Ollama version, generation hardware, and unspecified defaults are unavailable. Conclusions are conditional on the archived response pool.

For gate development, the 353 human-labeled sources are crossed with three roles and three model identities, yielding 3,177 response cells: 1,059 diagonal and 2,118 off-diagonal. Student supervision uses only the three diagonal responses per source. The 12,990 gate-candidate count and the 12,990 optimizer-row count introduced below are equal numerically but represent different units.

\section{Correctness-Gated Supervision}
\subsection{Composite candidate-correctness gate}
For candidate \(j\) associated with source \(i\), the gate estimates

\[
q_{ij}=P(z_{ij}=1\mid v_{ij},a_{ij},c_{ij},m_j,t_i),
\]

where \(z_{ij}=1\) requires structural validity and a decision equal to the adjudicated human training label. The feature vector contains 16 recorded primary features spanning verifier compatibility, diagonal consensus, teacher confidence, schema/task/citation validity, task, decision, and model identity. The primary no-role specification excludes role, task-role interaction, role-conditioned confidence, and role-conditioned verifier features, but retains model identity.

The gate uses scikit-learn LogisticRegression and StratifiedGroupKFold \citep{pedregosa2011scikit} with \(C=1\), balanced class weights, lbfgs, and 4,000 maximum iterations. Five shuffled folds (seed 260598) group by source\_example\_id. The 353 sources belong to 87 broader leakage\_group\_id clusters; 35 cross folds and 14 cross all five. This dependence can affect diagnostics, threshold selection, and supervision for the 353 human-source families. Thus \(q\) is an engineering score conditional on this split, not an out-of-cluster estimate.

The no-role gate has out-of-fold AUC 0.909101, Brier 0.119848, and log loss 0.379951. Threshold \(\tau=0.41\) maximizes pooled out-of-fold F1 over a 0.30--0.80 grid, with precision and then higher threshold as tie-breaks. There is no separate calibration stage, so \(q\) is not called a calibrated probability. Gate ablations and resampling details appear in Appendix B.

\subsection{Filtering, weighting, and eight supervision arms}
Let \(V_i\) contain structurally valid candidates with a legal task decision. The retained set is

\[
S_i=\{j\in V_i:q_{ij}\geq \tau\}.
\]

If this set is empty, the highest-scoring valid candidate is retained. The hard arm assigns equal within-family mass. The weighted arm uses

\[
w_{ij}=\frac{q_{ij}}{\sum_{k\in S_i}q_{ik}},\qquad j\in S_i.
\]

The 353 human source families use stored source-out-of-fold scores for their diagonal candidates. The other 3,977 families use the final gate fitted on all 3,177 gate-development response cells. The transported score has no independently labeled validation set in the weak/silver pool and is not evidence of calibration or correctness there. The hard and correctness-weighted arms retain the same candidate IDs.

Per virtual epoch, the hard and weighted arms retain 8,578/12,990 candidates: 8,257 pass \(\tau\), and 321 otherwise empty families use deterministic best-candidate fallback. There are no downstream rationale fallbacks.

Selected IDs and materialized decisions match for 4,330/4,330 families, whereas soft targets match for 4,324/4,330 and chosen conversations for 4,202/4,330. The arms are retained-set and decision matched, not fully target matched or reducible to a scalar-only perturbation. Appendix B reports the retained composition and decision counts.

The eight arms are source-label SFT, answer-only KD, single-teacher rationale KD, uniform unfiltered MTD, hard filtering, correctness-weighted MTD, dynamic-product MTD, and role-aware weighting. They provide direct-supervision, transfer, removal-stage, alternative-weight, and explicit-role-feature controls around the primary weighted-versus-unfiltered comparison.

This design supports a matched comparison of the implemented hard and weighted pipelines, but not a scalar-weight-only effect: relative masses, 6/4,330 soft targets, 128/4,330 serialized conversations, and their training interactions differ despite identical retained candidate IDs and materialized decisions.

\subsection{Student targets, prompt contract, and decoder}
The retained teacher mass defines a soft decision target. Claim targets factor into evidence sufficiency and polarity conditional on sufficiency:

\[
p_i^{\mathrm{suf}}=1-p_i(\mathrm{NotEnoughInfo}),\qquad
p_i^{\mathrm{sup}\mid\mathrm{suf}}=
\frac{p_i(\mathrm{Supported})}{p_i(\mathrm{Supported})+p_i(\mathrm{Refuted})}.
\]

Queries use \(p_i^{\mathrm{ans}}=p_i(\mathrm{Answer})\). Three binary heads predict sufficiency, conditional polarity, and answerability from the prompt-end hidden state. The causal language model generates the answer, rationale, and citations. The explicit Decision span is masked from language-model loss. For applicable task terms,

\[
\mathcal{L}=\mathcal{L}_{LM}+\mathcal{L}_{suf}
+p_i^{\mathrm{suf}}\mathcal{L}_{pol}+\mathcal{L}_{ans}.
\]

When retained sufficient mass is zero, the conditional-polarity target is set to 0.5 and its loss weight to 0, so that row contributes no polarity-head gradient. Sufficiency and polarity losses apply to claims, while answerability loss applies to queries.

The training system prompt instructs the model to abstain if evidence is insufficient, conflicting, or stale. The sealed-evaluation prompt omits the words or stale. Prompts are identical across arms within each phase, but training and evaluation prompts are not byte-identical across phases.

At inference, Answer requires probability at least 0.55 and a 0.10 margin over Abstain. A Supported or Refuted claim requires the same threshold and margin over NotEnoughInfo; otherwise the decoder emits NotEnoughInfo. The decision prefix is forced before common greedy generation. Full decoder settings appear in Appendix B.

\section{Experimental Design and Analysis}
\subsection{Controlled training matrix}
The student follows the dense MiniMind-3 architecture \citep{gong2026minimind}: hidden size 768, eight layers, 63,912,192 base parameters, and 2,307 decision-head parameters (63,914,499 total). Every arm has 12,990 physical rows, three virtual passes over 4,330 sources, 406 optimizer updates, and seeds 260631--260633.

Formal optimization uses BF16 on one NVIDIA GeForce RTX 3060 Ti; the saved checkpoints and sealed inference use FP16. All 24 arm-seed prediction files are available, whereas detailed run summaries cover the eight reference-seed runs. Appendix B gives optimizer and schedule settings.

The matrix matches architecture, training sources, physical rows, optimizer updates, evidence, and decoder. It is not token- or compute-matched across source-label, answer-only, and rationale objectives because target lengths and supervised language-model exposure differ. Contrasts across objective families therefore describe observed performance, not a pure effect of rationale availability.

\subsection{Outcomes and evidence tiers}
The three predeclared decision summaries are exact accuracy, unweighted five-label macro-F1, and the task-defined pooled conditional unsafe-action rate. They form a co-primary descriptive family, not a confirmatory multiple-testing family: no omnibus success decision is based on one interval or on their conjunction, and no multiplicity adjustment is claimed. Macro-F1 assigns zero F1 to a label with no predicted positives. The unsafe-action denominator contains 89 gold-NotEnoughInfo claims and 67 gold-Abstain queries, hence 156 safe-required cases in each observed run. Per-label recall and individual seed behavior form an anti-shortcut check. For each seed, automatic cross-encoder support, rationale, unsupported-material, and citation diagnostics are averaged over that system's emitted positives; positive denominators vary by arm and seed. Only response parse rate uses all outputs. These output-policy-conditional diagnostics are not paired grounding effects or human judgments.

The availability-amended human audit is a separate descriptive tier. It evaluates evidence support of the answer, rationale faithfulness, presence of an unsupported material claim, and citation validity among emitted positives at seed 260631. The manuscript reports raw counts because the audit is small and its system-specific positive samples are not paired.

Two legacy metric families, a pooled F1 field and a risk-coverage quantity, have unresolved cross-version estimand and provenance conflicts. Their values are excluded from every scientific result surface; Appendix F records this boundary.

\subsection{Statistical analysis}
The primary contrast is correctness-weighted minus unfiltered MTD. Its point estimand is the mean of the three seed-specific paired differences on the current held-out population. For each of 10,000 crossed-bootstrap replicates, the procedure samples the three observed training seeds with replacement and separately samples the 22 leakage groups with replacement. Within each sampled seed, the selected groups reconstruct paired candidate and baseline rows. The metric difference is computed from those rows and then averaged across sampled seeds. We report 2.5th and 97.5th percentiles as a 95\% observed-matrix percentile interval. The primary random-number seed is 260600. The six declared contrast jobs use fixed seeds 260598 through 260603 in their recorded order.

These intervals are descriptive stability summaries for the current three-seed by 22-group matrix. With only three training seeds, they are not calibrated confidence intervals for a future-seed population and should not be interpreted as uncertainty over future initializations, model families, or domains. Secondary arm comparisons are likewise descriptive and are not a multiplicity-adjusted confirmatory family. We say that an observed-matrix interval lies above or below zero, not that a population null has been rejected.

The archived decision for a positive grounded-distillation claim required more than directionally better aggregate endpoints. Human answer support and rationale faithfulness had to improve, unsupported material had to decrease, and every label needed nonzero recall in each seed. Automatic diagnostics could not satisfy human conditions. The present paper uses this rule only to bound interpretation, not to imply external preregistration.

\subsection{Availability-amended human audit}
The original audit requested 20 positive and 20 safe outputs per arm at seed 260631. After unsealing and automatic summarization, single-teacher rationale KD had only 14 safe outputs. Before human annotation, the procedure was amended to retain 20 positives per arm and cap safe rows at availability, yielding 160 positives and 154 safe rows. Appendix D and the accompanying source supplement preserve the timing, sampling, and rubric.

The positive stratum remains 20 outputs per arm. Source overlap was not serialized, so counts characterize each arm's realized positive policy rather than paired treatments; safe rows affect agreement only.

System, seed, gold label, and automatic scores were masked. Two annotators used Yes/No/Unclear, followed by disagreement-only adjudication. Unclear conservatively fails support, faithfulness, and citation validity and counts as unsupported material. Agreement statistics and field definitions are in Appendix D.

Figure~\ref{fig:1} summarizes the fixed processing order and separates its evidence channels. The post-summary, one-seed, non-paired audit is descriptive: it cannot estimate a common-source contrast or establish system-level improvement or harm, so raw counts are primary.

\begin{figure}[!htbp]
\centering
\includegraphics[width=0.98\linewidth]{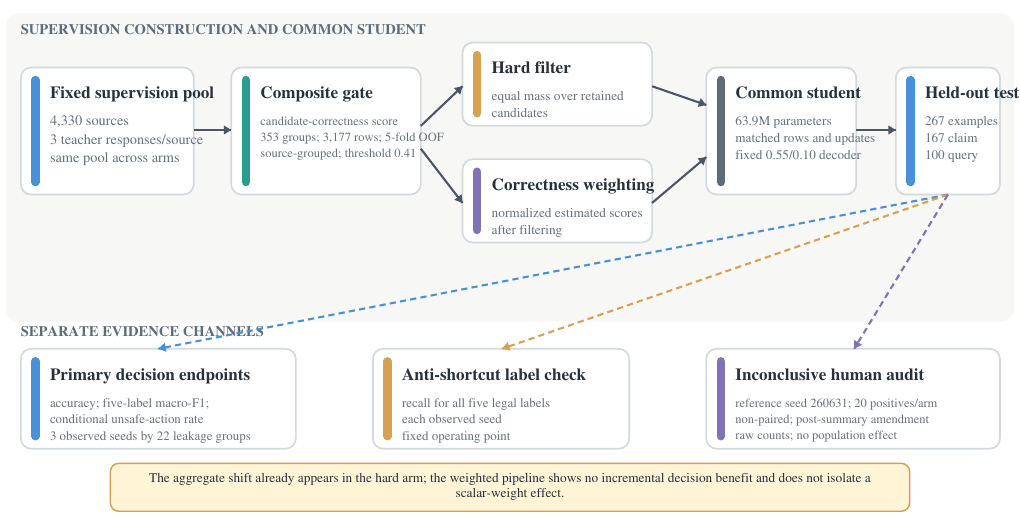}
\caption{Study design and inferential boundaries. A fixed three-response teacher pool feeds a composite candidate-correctness gate. The hard arm retains candidates at the fixed threshold and assigns equal within-family mass. The weighted arm uses the same retained IDs and assigns estimated-score mass. A common student and decoder are evaluated through three non-interchangeable channels: primary automatic decisions, per-label anti-shortcut checks, and an availability-amended descriptive human audit. The hard arm already exhibits the aggregate pattern, while the implemented weighted arm shows no demonstrated incremental decision benefit. Because target text is not fully matched, the comparison neither isolates a scalar-weight effect nor identifies a mediator.}
\label{fig:1}
\end{figure}

\section{Results}
\subsection{Primary decision contrast}
The correctness-gated weighted arm differed from unfiltered MTD on all three fixed decision summaries in the observed matrix. Accuracy was 0.634 rather than 0.468, five-label macro-F1 was 0.477 rather than 0.345, and the conditional unsafe-action rate was 0.135 rather than 0.632. The paired differences and observed-matrix intervals appear in Table~\ref{tab:2}.

These endpoint differences establish a change in decisions under the fixed student, teacher pool, test set, and decoder. They do not establish a change in the evidential quality of generated text. The distinction is already visible in the primary system summaries: the lower conditional unsafe-action rate is accompanied by substantially different label use, while the human positive-output sample contains no evidence-supported weighted answer.

Table~\ref{tab:1} gives the complete arm matrix. Values are seed means with sample standard deviations across the three observed training seeds. Source-label SFT has the highest five-label macro-F1, 0.586, whereas the hard-filter arm has the highest observed accuracy, 0.660. Hard filtering and correctness weighting tie at the displayed minimum conditional unsafe-action rate of 0.135. The matrix therefore does not support a single method winner.

\begin{table}[!htbp]
\caption{Automatic decision endpoints for all eight arms, mean \(\pm\) SD across three observed seeds.}
\label{tab:1}
\centering
\small
\setlength{\tabcolsep}{3.5pt}
\renewcommand{\arraystretch}{1.12}
\begin{tabularx}{\linewidth}{@{}>{\raggedright\arraybackslash}p{0.12\linewidth}>{\raggedright\arraybackslash}X>{\raggedright\arraybackslash}X>{\raggedright\arraybackslash}X@{}}
\toprule
\textbf{Arm} & \textbf{Accuracy \(\uparrow\)} & \textbf{Five-label macro-F1 \(\uparrow\)} & \textbf{Task-defined conditional unsafe-action rate \(\downarrow\)} \\
\midrule
Source-label SFT & \(0.588 \pm 0.021\) & \textbf{\(0.586 \pm 0.012\)} & \(0.434 \pm 0.052\) \\
Answer-only KD & \(0.353 \pm 0.022\) & \(0.308 \pm 0.014\) & \(0.983 \pm 0.015\) \\
Single-teacher rationale KD & \(0.380 \pm 0.028\) & \(0.340 \pm 0.032\) & \(0.938 \pm 0.033\) \\
Unfiltered MTD & \(0.468 \pm 0.010\) & \(0.345 \pm 0.006\) & \(0.632 \pm 0.026\) \\
Correctness hard filter & \textbf{\(0.660 \pm 0.018\)} & \(0.530 \pm 0.073\) & \textbf{\(0.135 \pm 0.039\)} \\
Correctness-weighted MTD & \(0.634 \pm 0.017\) & \(0.477 \pm 0.014\) & \textbf{\(0.135 \pm 0.044\)} \\
Dynamic-product MTD & \(0.637 \pm 0.006\) & \(0.489 \pm 0.026\) & \(0.139 \pm 0.013\) \\
Role-aware correctness-weighted MTD & \(0.634 \pm 0.008\) & \(0.480 \pm 0.015\) & \(0.137 \pm 0.027\) \\
\bottomrule
\end{tabularx}
\end{table}

Source-label SFT is a particularly important control. The correctness-gated weighted arm has higher observed accuracy and a lower narrow unsafe-action rate, but its five-label macro-F1 is 0.109 lower. The ladder is also non-monotonic. Relative to SFT, single-teacher rationale KD differs in macro-F1 by \(-0.2460\), with a 95\% observed-matrix interval \([-0.3386,-0.1889]\), and in conditional unsafe action by \(+0.5043\), \([0.3279,0.6578]\). Unfiltered MTD then differs from single-teacher rationale KD in accuracy by \(+0.0886\), \([0.0249,0.1803]\), and unsafe action by \(-0.3056\), \([-0.4137,-0.2375]\), but it does not recover the SFT macro-F1. Rationale-bearing supervision is not monotonically beneficial in this matrix.

\subsection{Primary and component contrasts}
Table~\ref{tab:2} separates the primary arm comparison from descriptive component contrasts. The correctness-gated weighted arm minus unfiltered MTD has an accuracy difference of \(+0.1660\), with a 95\% observed-matrix interval \([0.0670,0.2455]\), a macro-F1 difference of \(+0.1323\), \([0.0916,0.1731]\), and a conditional unsafe-action difference of \(-0.4979\), \([-0.5926,-0.3686]\). The intervals are above zero for accuracy and macro-F1 and below zero for the narrowly defined unsafe-action endpoint; these directions do not imply preserved label functionality or grounding quality.

The hard-filter comparison changes the interpretation. Hard filtering minus unfiltered MTD is \(+0.1923\) for accuracy, \(+0.1856\) for macro-F1, and \(-0.4979\) for conditional unsafe action. The aggregate pattern is already present in the hard-filter arm before continuous weighting is added. The weighted arm is lower than the retained-candidate-set-matched hard arm by \(-0.0262\) in accuracy and \(-0.0533\) in macro-F1, with an unsafe-action difference of 0. The accuracy interval lies below zero. The macro-F1 and unsafe-action intervals include zero. The implemented weighted arm shows no demonstrated incremental decision benefit over hard filtering. As Section 4.2 records, the arms match all selected IDs and decisions but differ for 6/4,330 soft targets and 128/4,330 conversations, so this is not a fully target-matched scalar-weight experiment.

\begin{table}[!htbp]
\caption{Arm and component contrasts with 95\% percentile intervals over the observed three-seed by 22-group matrix. All intervals are descriptive and not multiplicity adjusted.}
\label{tab:2}
\centering
\small
\setlength{\tabcolsep}{3.5pt}
\renewcommand{\arraystretch}{1.12}
\resizebox{\linewidth}{!}{%
\begin{tabular}{@{}lcccc@{}}
\toprule
\textbf{Contrast} & \textbf{Evidence tier} & \textbf{Accuracy} & \textbf{Five-label macro-F1} & \textbf{Conditional unsafe-action rate} \\
\midrule
Correctness weighted \(-\) unfiltered & Primary decision & \(+0.1660\ [0.0670,0.2455]\) & \(+0.1323\ [0.0916,0.1731]\) & \(-0.4979\ [-0.5926,-0.3686]\) \\
Hard filter \(-\) unfiltered & Descriptive component & \(+0.1923\ [0.1075,0.2641]\) & \(+0.1856\ [0.1157,0.2764]\) & \(-0.4979\ [-0.5862,-0.4007]\) \\
Correctness weighted \(-\) hard filter & Descriptive component & \(-0.0262\ [-0.0597,-0.0024]\) & \(-0.0533\ [-0.1377,0.0088]\) & \(0.0000\ [-0.0710,0.0764]\) \\
Role-aware \(-\) no-role weighted & Secondary ablation & \(0.0000\ [-0.0124,0.0248]\) & \(+0.0033\ [-0.0163,0.0385]\) & \(+0.0021\ [-0.0328,0.0248]\) \\
\bottomrule
\end{tabular}%
}
\end{table}

The hard comparison is not a causal mediation estimate. Although the hard and weighted arms retain the same IDs and have the same materialized decisions, within-family masses and occasional target-text differences can alter learning dynamics. The comparison rules out a demonstrated incremental benefit of the implemented continuous-weighting arm in this experiment. It does not prove that candidate removal is necessary, sufficient, or uniquely responsible.

\subsection{Per-label behavior at the fixed operating point}
\begin{figure}[!htbp]
\centering
\includegraphics[width=0.98\linewidth]{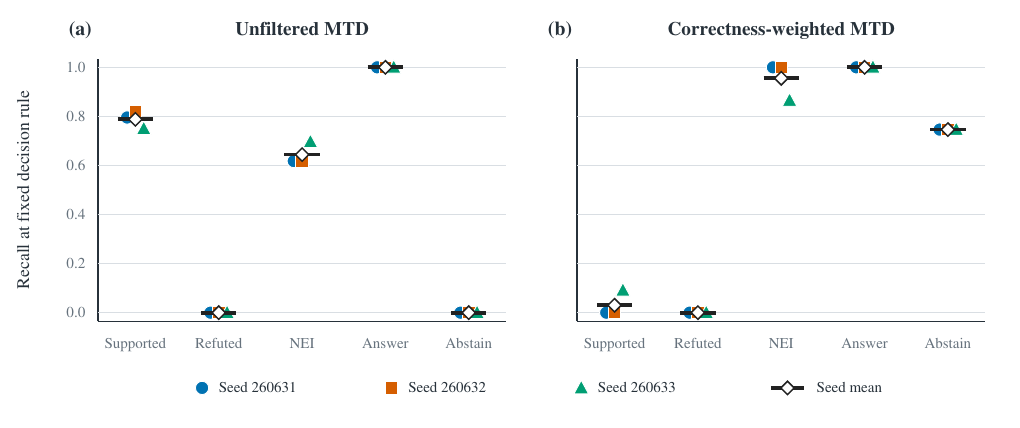}
\caption{Per-label recall at the fixed operating point. Points show each of the three observed training seeds and bars show seed means for unfiltered and correctness-weighted MTD. The figure uses the common 0.55 positive threshold and 0.10 margin. Zero Refuted recall in all three seeds, and no positive claim emissions in two seeds, are operating-point results rather than representation-level claims.}
\label{fig:2}
\end{figure}

Figure~\ref{fig:2} shows why the lower conditional unsafe-action rate cannot be read as general safety. Unfiltered MTD has mean recall 0.788 for Supported, 0 for Refuted, 0.644 for NotEnoughInfo, 1.000 for Answer, and 0 for Abstain. Correctness-weighted MTD has corresponding recall 0.030, 0, 0.955, 1.000, and 0.746.

The weighted arm therefore recovers Abstain and substantially improves NotEnoughInfo recall while nearly eliminating positive claim decisions. Seeds 260631 and 260632 assign NotEnoughInfo to all 167 claim examples. Their largest positive claim probabilities are 0.542 and 0.545, just below the fixed 0.55 threshold. Seed 260633 emits some Supported predictions, with Supported recall 0.091, but Refuted recall remains zero. No Refuted example is correctly recovered at the fixed operating point in any weighted seed; this does not mean that the label is absent from the held-out set.

This loss of label functionality is threshold-sensitive. The proximity of two seeds to 0.55 does not show that their hidden states contain no polarity information. It does show that the declared decoder produces no positive claim functionality for those runs. Retuning the decoder after opening the held-out test would change the study and is not performed.

\subsection{Inconclusive human grounding audit}
The reference-seed human audit does not establish improved grounding (Table~\ref{tab:3}). Among 20 unfiltered positive outputs, 1 is evidence-supported, 5 have faithful rationales, 19 contain unsupported material, and 1 has a valid citation. Among 20 weighted positives, the corresponding counts are 0, 11, 20, and 0. Higher observed rationale faithfulness in the weighted sample coexists with no supported answer, unsupported material in every sampled output, and no valid citation.

The full eight-arm view reinforces the boundary. Source-label SFT, which has the highest mean five-label macro-F1, also has the highest sampled answer-support count (7/20); Answer-only KD has the highest sampled valid-citation count (8/20). Yet 13/20 sampled SFT positives contain unsupported material. No arm is a validated high-grounding positive control. These counts describe realized positive-output policies at one seed. They neither rank system-level grounding nor estimate an average treatment effect across common examples.

\begin{table}[!htbp]
\caption{System-label-masked, availability-amended human audit among emitted positives at reference seed 260631. Each arm contributes 20 non-paired positive outputs. Raw counts are primary.}
\label{tab:3}
\centering
\small
\setlength{\tabcolsep}{3.5pt}
\renewcommand{\arraystretch}{1.12}
\resizebox{\linewidth}{!}{%
\begin{tabular}{@{}lcccc@{}}
\toprule
\textbf{Arm} & \textbf{Evidence-supported answer} & \textbf{Faithful rationale} & \textbf{Contains unsupported material \(\downarrow\)} & \textbf{Valid citation} \\
\midrule
Source-label SFT & 7/20 & 0/20 & 13/20 & 7/20 \\
Answer-only KD & 4/20 & 0/20 & 12/20 & 8/20 \\
Single-teacher rationale KD & 1/20 & 3/20 & 14/20 & 6/20 \\
Unfiltered MTD & 1/20 & 5/20 & 19/20 & 1/20 \\
Correctness hard filter & 4/20 & 4/20 & 16/20 & 4/20 \\
Correctness-weighted MTD & 0/20 & 11/20 & 20/20 & 0/20 \\
Dynamic-product MTD & 0/20 & 9/20 & 20/20 & 0/20 \\
Role-aware correctness-weighted MTD & 0/20 & 11/20 & 20/20 & 0/20 \\
\bottomrule
\end{tabular}%
}
\end{table}

The audit was amended after outputs became available but before annotation, used one reference seed, and did not serialize cross-arm source overlap. The strongest defensible statement is that the sampled weighted positives did not meet the stated answer-support and unsupported-material criteria. Because the samples are system-specific and non-paired, the audit is inconclusive about a population-level grounding benefit or harm.

Secondary automatic, explicit-role-feature, and timing diagnostics appear in Appendix B; none supplies human grounding evidence.

\section{Discussion}
The weighted arm shifts decisions but loses label functionality: no audited answer is supported, all contain unsupported material, Refuted recall is zero, and two seeds emit no positive claim. The non-paired audit precludes a grounding effect claim in either direction. Decision correctness does not supervise evidence-licensed language, but objective mismatch is only one possible mechanism. The hard arm already shows the pattern; weighted changes mass and 128/4,330 conversations despite matched IDs and decisions, so causal isolation requires independent manipulation. Grounding-aware critics remain untested \citep{kawabata2024rationaleaware,gao2025critics}. Selectivity and calibration findings \citep{xin2021abstention,geifman2017selective,kamath2020selective,guo2017calibration} require reporting coverage, class functionality, and evidence quality together.

\section{Limitations}
One student, domain, teacher pool, decoder, 267 examples, and three seeds preclude population or scaling claims. Broader-group dependence, weak/silver provenance, runtime/phase gaps, threshold sensitivity, and the amended non-paired audit further limit causal and retraining claims (Appendices B, D, and E).

\section{Ethics and Responsible Research}
Restaurant reviews can contain personal information and consequential allegations; restricted text is separated, but unsupported or stale claims remain harmful. This non-deployment study's narrow unsafe metric omits harmful language, fairness, privacy, retrieval error, and downstream consequences.

Humans created the 882-example corpus and reviewed the 314-row audit, yet recruitment, compensation, consent, privacy, and institutional determination are unrecorded; we make no claim, and this blocks submission.

\section{Reproducibility and Data Availability}
An internal record of configurations, manifests, implementations, 24 sealed prediction files with held-out labels, masked audit materials, and checksums reproduced every arm-seed summary and contrast within \(10^{-12}\), but not training (Appendix E).

The arXiv source package includes source, figures, bibliography, aggregate audit, and protocols; it excludes predictions, Yelp text, labels, raw reviews, identity mappings, and weights. It supports aggregate inspection, not row-level recomputation; repository/DOI, access, and permissions remain unverified.

\section{Conclusion}
The weighted arm shifts decisions but loses label functionality: SFT leads macro-F1; Refuted recall is zero; two seeds emit no positive claim; and its human sample has 0/20 supported answers and 20/20 positives with unsupported material. The hard arm already exhibits the pattern, with no weighted increment. This is decision redistribution, not grounding improvement or causal identification.

\bibliography{references}

@inproceedings{hsieh2023distilling,
  author    = {Cheng-Yu Hsieh and Chun-Liang Li and Chih-kuan Yeh and Hootan Nakhost and Yasuhisa Fujii and Alex Ratner and Ranjay Krishna and Chen-Yu Lee and Tomas Pfister},
  title     = {Distilling Step-by-Step! Outperforming Larger Language Models with Less Training Data and Smaller Model Sizes},
  booktitle = {Findings of the Association for Computational Linguistics: ACL 2023},
  year      = {2023},
  pages     = {8003--8017},
  doi       = {10.18653/v1/2023.findings-acl.507}
}

@inproceedings{shridhar2023reasoning,
  author    = {Kumar Shridhar and Alessandro Stolfo and Mrinmaya Sachan},
  title     = {Distilling Reasoning Capabilities into Smaller Language Models},
  booktitle = {Findings of the Association for Computational Linguistics: ACL 2023},
  year      = {2023},
  pages     = {7059--7073},
  doi       = {10.18653/v1/2023.findings-acl.441}
}

@inproceedings{wang2023scott,
  author    = {Peifeng Wang and Zhengyang Wang and Zheng Li and Yifan Gao and Bing Yin and Xiang Ren},
  title     = {{SCOTT}: Self-Consistent Chain-of-Thought Distillation},
  booktitle = {Proceedings of the 61st Annual Meeting of the Association for Computational Linguistics (Volume 1: Long Papers)},
  year      = {2023},
  pages     = {5546--5558},
  doi       = {10.18653/v1/2023.acl-long.304}
}

@inproceedings{lee2024mentorkd,
  author    = {Hojae Lee and Junho Kim and SangKeun Lee},
  title     = {{Mentor-KD}: Making Small Language Models Better Multi-step Reasoners},
  booktitle = {Proceedings of the 2024 Conference on Empirical Methods in Natural Language Processing},
  year      = {2024},
  pages     = {17643--17658},
  doi       = {10.18653/v1/2024.emnlp-main.977}
}

@inproceedings{tian2025beyond,
  author    = {Yijun Tian and Yikun Han and Xiusi Chen and Wei Wang and Nitesh V. Chawla},
  title     = {Beyond Answers: Transferring Reasoning Capabilities to Smaller {LLM}s Using Multi-Teacher Knowledge Distillation},
  booktitle = {Proceedings of the Eighteenth ACM International Conference on Web Search and Data Mining},
  year      = {2025},
  pages     = {251--260},
  doi       = {10.1145/3701551.3703577}
}

@article{yuan2021reinforced,
  author  = {Fei Yuan and Linjun Shou and Jian Pei and Wutao Lin and Ming Gong and Yan Fu and Daxin Jiang},
  title   = {Reinforced Multi-Teacher Selection for Knowledge Distillation},
  journal = {Proceedings of the AAAI Conference on Artificial Intelligence},
  year    = {2021},
  volume  = {35},
  number  = {16},
  pages   = {14284--14291},
  doi     = {10.1609/aaai.v35i16.17680}
}

@article{ding2024tradeoff,
  author  = {Zixiang Ding and Guoqing Jiang and Shuai Zhang and Lin Guo and Wei Lin},
  title   = {How to Trade Off the Quantity and Capacity of Teacher Ensemble: Learning Categorical Distribution to Stochastically Employ a Teacher for Distillation},
  journal = {Proceedings of the AAAI Conference on Artificial Intelligence},
  year    = {2024},
  volume  = {38},
  number  = {16},
  pages   = {17915--17923},
  doi     = {10.1609/aaai.v38i16.29746}
}

@inproceedings{li2025committee,
  author    = {Zhuochun Li and Yuelyu Ji and Rui Meng and Daqing He},
  title     = {Learning from Committee: Reasoning Distillation from a Mixture of Teachers with Peer-Review},
  booktitle = {Findings of the Association for Computational Linguistics: ACL 2025},
  year      = {2025},
  pages     = {4190--4205},
  doi       = {10.18653/v1/2025.findings-acl.217}
}

@inproceedings{xin2021abstention,
  author    = {Ji Xin and Raphael Tang and Yaoliang Yu and Jimmy Lin},
  title     = {The Art of Abstention: Selective Prediction and Error Regularization for Natural Language Processing},
  booktitle = {Proceedings of the 59th Annual Meeting of the Association for Computational Linguistics and the 11th International Joint Conference on Natural Language Processing (Volume 1: Long Papers)},
  year      = {2021},
  pages     = {1040--1051},
  doi       = {10.18653/v1/2021.acl-long.84}
}

@inproceedings{feng2024abstain,
  author    = {Shangbin Feng and Weijia Shi and Yike Wang and Wenxuan Ding and Orevaoghene Ahia and Shuyue Stella Li and Vidhisha Balachandran and Sunayana Sitaram and Yulia Tsvetkov},
  title     = {Teaching {LLM}s to Abstain across Languages via Multilingual Feedback},
  booktitle = {Proceedings of the 2024 Conference on Empirical Methods in Natural Language Processing},
  year      = {2024},
  pages     = {4125--4150},
  doi       = {10.18653/v1/2024.emnlp-main.239}
}

@inproceedings{madhusudhan2025notanswer,
  author    = {Nishanth Madhusudhan and Sathwik Tejaswi Madhusudhan and Vikas Yadav and Masoud Hashemi},
  title     = {Do {LLM}s Know When to {NOT} Answer? Investigating Abstention Abilities of Large Language Models},
  booktitle = {Proceedings of the 31st International Conference on Computational Linguistics},
  year      = {2025},
  pages     = {9329--9345},
  url       = {https://aclanthology.org/2025.coling-main.627/}
}

@inproceedings{gangrade2021selective,
  author    = {Aditya Gangrade and Anil Kag and Venkatesh Saligrama},
  title     = {Selective Classification via One-Sided Prediction},
  booktitle = {Proceedings of the 24th International Conference on Artificial Intelligence and Statistics},
  year      = {2021},
  series    = {Proceedings of Machine Learning Research},
  volume    = {130},
  pages     = {2179--2187},
  url       = {https://proceedings.mlr.press/v130/gangrade21a.html}
}

@inproceedings{wang2025qcrd,
  author    = {Wei Wang and Zhaowei Li and Qi Xu and YiQing Cai and Hang Song and Qi Qi and Ran Zhou and Zhida Huang and Tao Wang and Li Xiao},
  title     = {{QCRD}: Quality-guided Contrastive Rationale Distillation for Large Language Models},
  booktitle = {Proceedings of the 2025 Conference on Empirical Methods in Natural Language Processing},
  year      = {2025},
  pages     = {14334--14345},
  doi       = {10.18653/v1/2025.emnlp-main.724}
}

@inproceedings{zhang2024elad,
  author    = {Yifei Zhang and Bo Pan and Chen Ling and Yuntong Hu and Liang Zhao},
  title     = {{ELAD}: Explanation-Guided Large Language Models Active Distillation},
  booktitle = {Findings of the Association for Computational Linguistics: ACL 2024},
  year      = {2024},
  pages     = {4463--4475},
  doi       = {10.18653/v1/2024.findings-acl.264}
}

@inproceedings{yan2025efficient,
  author    = {JianZhi Yan and Le Liu and Youcheng Pan and Shiwei Chen and Yang Xiang and Buzhou Tang},
  title     = {Towards Efficient {CoT} Distillation: Self-Guided Rationale Selector for Better Performance with Fewer Rationales},
  booktitle = {Findings of the Association for Computational Linguistics: EMNLP 2025},
  year      = {2025},
  pages     = {7818--7835},
  doi       = {10.18653/v1/2025.findings-emnlp.413}
}

@inproceedings{song2025rationale,
  author    = {Hoyun Song and Huije Lee and Jisu Shin and Sukmin Cho and Changgeon Ko and Jong C. Park},
  title     = {Does Rationale Quality Matter? Enhancing Mental Disorder Detection via Selective Reasoning Distillation},
  booktitle = {Findings of the Association for Computational Linguistics: ACL 2025},
  year      = {2025},
  pages     = {21738--21756},
  doi       = {10.18653/v1/2025.findings-acl.1119}
}

@article{liu2023selective,
  author  = {Min Liu and Yu Bao and Chengqi Zhao and Shujian Huang},
  title   = {Selective Knowledge Distillation for Non-Autoregressive Neural Machine Translation},
  journal = {Proceedings of the AAAI Conference on Artificial Intelligence},
  year    = {2023},
  volume  = {37},
  number  = {11},
  pages   = {13246--13254},
  doi     = {10.1609/aaai.v37i11.26555}
}

@misc{hosseini2024vstar,
  author        = {Arian Hosseini and Xingdi Yuan and Nikolay Malkin and Aaron Courville and Alessandro Sordoni and Rishabh Agarwal},
  title         = {{V-STaR}: Training Verifiers for Self-Taught Reasoners},
  year          = {2024},
  howpublished  = {arXiv},
  eprint        = {2402.06457},
  archiveprefix = {arXiv},
  url           = {https://arxiv.org/abs/2402.06457}
}

@inproceedings{kawabata2024rationaleaware,
  author    = {Akira Kawabata and Saku Sugawara},
  title     = {Rationale-Aware Answer Verification by Pairwise Self-Evaluation},
  booktitle = {Proceedings of the 2024 Conference on Empirical Methods in Natural Language Processing},
  year      = {2024},
  pages     = {16178--16196},
  doi       = {10.18653/v1/2024.emnlp-main.905}
}

@inproceedings{zhang2024strongverifiers,
  author    = {Yunxiang Zhang and Muhammad Khalifa and Lajanugen Logeswaran and Jaekyeom Kim and Moontae Lee and Honglak Lee and Lu Wang},
  title     = {Small Language Models Need Strong Verifiers to Self-Correct Reasoning},
  booktitle = {Findings of the Association for Computational Linguistics: ACL 2024},
  year      = {2024},
  pages     = {15637--15653},
  doi       = {10.18653/v1/2024.findings-acl.924}
}

@misc{dixit2026aletheia,
  author        = {Aradhya Dixit and Tianxi Liang and Jai Telang},
  title         = {Project Aletheia: Verifier-Guided Distillation of Backtracking for Small Language Models},
  year          = {2026},
  howpublished  = {arXiv},
  eprint        = {2601.14290},
  archiveprefix = {arXiv},
  url           = {https://arxiv.org/abs/2601.14290}
}

@inproceedings{wang2024rdrec,
  author    = {Xinfeng Wang and Jin Cui and Yoshimi Suzuki and Fumiyo Fukumoto},
  title     = {{RDRec}: Rationale Distillation for {LLM}-based Recommendation},
  booktitle = {Proceedings of the 62nd Annual Meeting of the Association for Computational Linguistics (Volume 2: Short Papers)},
  year      = {2024},
  pages     = {65--74},
  doi       = {10.18653/v1/2024.acl-short.6}
}

@inproceedings{wu2021oneteacher,
  author    = {Chuhan Wu and Fangzhao Wu and Yongfeng Huang},
  title     = {One Teacher is Enough? Pre-trained Language Model Distillation from Multiple Teachers},
  booktitle = {Findings of the Association for Computational Linguistics: ACL-IJCNLP 2021},
  year      = {2021},
  pages     = {4408--4413},
  doi       = {10.18653/v1/2021.findings-acl.387}
}

@inproceedings{gao2025critics,
  author    = {Bofei Gao and Zefan Cai and Runxin Xu and Peiyi Wang and Ce Zheng and Runji Lin and Keming Lu and Dayiheng Liu and Chang Zhou and Wen Xiao and Tianyu Liu and Baobao Chang},
  title     = {{LLM} Critics Help Catch Bugs in Mathematics: Towards a Better Mathematical Verifier with Natural Language Feedback},
  booktitle = {Findings of the Association for Computational Linguistics: ACL 2025},
  year      = {2025},
  pages     = {14588--14604},
  doi       = {10.18653/v1/2025.findings-acl.753}
}

@article{atanasova2022insufficient,
  author  = {Pepa Atanasova and Jakob Grue Simonsen and Christina Lioma and Isabelle Augenstein},
  title   = {Fact Checking with Insufficient Evidence},
  journal = {Transactions of the Association for Computational Linguistics},
  year    = {2022},
  volume  = {10},
  pages   = {746--763},
  doi     = {10.1162/tacl_a_00486}
}

@inproceedings{kamath2020selective,
  author    = {Amita Kamath and Robin Jia and Percy Liang},
  title     = {Selective Question Answering under Domain Shift},
  booktitle = {Proceedings of the 58th Annual Meeting of the Association for Computational Linguistics},
  year      = {2020},
  pages     = {5684--5696},
  doi       = {10.18653/v1/2020.acl-main.503}
}

@inproceedings{tayebati2025cap,
  author    = {Sina Tayebati and Divake Kumar and Nastaran Darabi and Dinithi Jayasuriya and Theja Tulabandhula and Ranganath Krishnan and Amit Ranjan Trivedi},
  title     = {{CAP}: Conformalized Abstention Policies for Context-Adaptive Risk Management for {LLM}s and {VLM}s},
  booktitle = {Proceedings of the 17th Asian Conference on Machine Learning},
  year      = {2025},
  series    = {Proceedings of Machine Learning Research},
  volume    = {304},
  pages     = {926--941},
  url       = {https://proceedings.mlr.press/v304/tayebati26a.html}
}

@misc{hinton2015distilling,
  author        = {Geoffrey Hinton and Oriol Vinyals and Jeff Dean},
  title         = {Distilling the Knowledge in a Neural Network},
  year          = {2015},
  howpublished  = {arXiv},
  eprint        = {1503.02531},
  archiveprefix = {arXiv},
  url           = {https://arxiv.org/abs/1503.02531}
}

@inproceedings{geifman2017selective,
  author    = {Yonatan Geifman and Ran El-Yaniv},
  title     = {Selective Classification for Deep Neural Networks},
  booktitle = {Advances in Neural Information Processing Systems},
  year      = {2017},
  volume    = {30},
  pages     = {4878--4887},
  url       = {https://proceedings.neurips.cc/paper_files/paper/2017/hash/4a8423d5e91fda00bb7e46540e2b0cf1-Abstract.html}
}

@inproceedings{guo2017calibration,
  author    = {Chuan Guo and Geoff Pleiss and Yu Sun and Kilian Q. Weinberger},
  title     = {On Calibration of Modern Neural Networks},
  booktitle = {Proceedings of the 34th International Conference on Machine Learning},
  year      = {2017},
  series    = {Proceedings of Machine Learning Research},
  volume    = {70},
  pages     = {1321--1330},
  url       = {https://proceedings.mlr.press/v70/guo17a.html}
}

@misc{yelp2026opendataset,
  author       = {{Yelp}},
  title        = {Yelp Open Dataset},
  year         = {2026},
  howpublished = {Official data-licensing website},
  url          = {https://business.yelp.com/data/resources/open-dataset/},
  note         = {Accessed 30 August 2026}
}

@misc{gong2026minimind,
  author       = {Jingyao Gong and contributors},
  title        = {{MiniMind}: Training a 64M-Parameter Language Model from Scratch},
  year         = {2026},
  howpublished = {Official project repository},
  url          = {https://github.com/jingyaogong/minimind},
  note         = {MiniMind-3 release documentation; accessed 30 August 2026}
}

@misc{ollama2026command,
  author       = {{Ollama}},
  title        = {{command-r7b} Model Library Entry},
  year         = {2026},
  howpublished = {Official Ollama model library},
  url          = {https://ollama.com/library/command-r7b},
  note         = {Accessed 30 August 2026}
}

@misc{ollama2026qwen,
  author       = {{Ollama}},
  title        = {{qwen3.5} Model Tags},
  year         = {2026},
  howpublished = {Official Ollama model library},
  url          = {https://ollama.com/library/qwen3.5/tags},
  note         = {Accessed 30 August 2026}
}

@misc{ollama2026granite,
  author       = {{Ollama}},
  title        = {{granite4.1} Model Tags},
  year         = {2026},
  howpublished = {Official Ollama model library},
  url          = {https://ollama.com/library/granite4.1/tags},
  note         = {Accessed 30 August 2026}
}

@article{pedregosa2011scikit,
  author  = {Fabian Pedregosa and Gael Varoquaux and Alexandre Gramfort and Vincent Michel and Bertrand Thirion and Olivier Grisel and Mathieu Blondel and Peter Prettenhofer and Ron Weiss and Vincent Dubourg and Jake VanderPlas and Alexandre Passos and David Cournapeau and Matthieu Brucher and Matthieu Perrot and {\'E}douard Duchesnay},
  title   = {Scikit-learn: Machine Learning in Python},
  journal = {Journal of Machine Learning Research},
  year    = {2011},
  volume  = {12},
  pages   = {2825--2830},
  url     = {https://jmlr.org/papers/v12/pedregosa11a.html}
}
\bibliographystyle{plainnat}
\appendix
\setcounter{table}{0}
\renewcommand{\thetable}{A\arabic{table}}
\renewcommand{\theHtable}{A.\arabic{table}}

\section{Study Units and Label Distributions}
The active human-gold splits are fully separated from the weak/silver provenance of the larger training pool. Table~\ref{tab:a1} records the five-label counts used to interpret class-specific behavior.

\begin{table}[!htbp]
\caption{Human-gold label distributions.}
\label{tab:a1}
\centering
\small
\setlength{\tabcolsep}{3.5pt}
\renewcommand{\arraystretch}{1.12}
\resizebox{\linewidth}{!}{%
\begin{tabular}{@{}lrrrrrr@{}}
\toprule
\textbf{Split} & \textbf{Supported} & \textbf{Refuted} & \textbf{NotEnoughInfo} & \textbf{Answer} & \textbf{Abstain} & \textbf{Total} \\
\midrule
Train & 83 & 53 & 117 & 34 & 66 & 353 \\
Legacy development & 40 & 30 & 92 & 31 & 69 & 262 \\
Held-out test & 44 & 34 & 89 & 33 & 67 & 267 \\
\bottomrule
\end{tabular}%
}
\end{table}

The 4,330-source student pool contains 353 human-gold and 3,977 weak/silver sources. Its task counts are 3,598 claim and 732 query sources. The source reference-label totals are 1,294 Supported, 1,122 Refuted, 1,182 NotEnoughInfo, 348 Answer, and 384 Abstain.

\section{Gate, Teacher, and Prompt Diagnostics}
Teacher generation uses recommendation/command-r7b:latest, verification/qwen3.5:9b-q4\_K\_M, and abstention/granite4.1:8b-q5\_K\_M. Their seeds are 5981, 5982, and 5983; temperatures are 0.2, 0, and 0; each has a 768-token cap and at most two attempts. Requests run in role-then-request order with strict task-restricted JSON. Serialized messages include task, claim or query, evidence and identifiers, time bin, entity, role, evidence-only instruction, and length constraints. Runtime top-p/top-k defaults were not recorded.

The primary gate uses 16 no-role features and five source-grouped folds. Removing verifier features raises Brier score from 0.119848 to 0.187486. Adding explicit role terms changes Brier by \(-0.000119\), with a 95\% percentile interval \([-0.000647,0.000408]\) from 10,000 paired source-group bootstrap replicates over 353 source\_example\_id groups (seed 260598); broader leakage groups remain unprotected. The selected no-role model is refit on all 3,177 gate-development cells. Its application to 3,977 weak/silver families is transported without independent labeled validation.

The 8,578 retained candidates comprise 8,257 threshold passes and 321 best-candidate fallbacks, all in claim families (227 Refuted, 68 Supported, 26 NotEnoughInfo reference labels). Families retain one/two/three candidates in counts 1,416/1,580/1,334. Retained roles number 3,306 verification, 2,638 abstention, and 2,634 recommendation. Hard and weighted materializations have zero rationale fallback and identical decision totals: 1,292 Supported, 815 Refuted, 1,491 NotEnoughInfo, 249 Answer, and 483 Abstain.

Greedy generation forces the selected decision prefix and uses at most 256 new tokens, temperature 1, top-p 1, top-k 0, and repetition penalty 1. The training prompt says insufficient, conflicting, or stale; evaluation omits or stale. Arms are prompt matched within, not across, phases.

AdamW uses microbatch 4, accumulation 8 (effective batch 32), base/head learning rates \(10^{-5}/10^{-4}\), weight decay 0.01, clipping 1, prompt/sequence limits 640/1,024, and four workers. A cosine schedule decays to a 0.1 terminal multiplier without serialized warmup. Only the final 406th update is retained. At seed 260631, elapsed time is 7.00 minutes unfiltered and 6.98 weighted. Across 801 predictions per arm, latency is 639.7 versus 672.9 ms and generated length 102.6 versus 108.4 tokens; these exclude preprocessing and gate/teacher costs.

Automatic support is 0.304 unfiltered and 0.034 weighted, conditioned on each arm's emitted positives. Explicit-role minus no-role weighted changes student macro-F1 by \(+0.0033\), accuracy by 0, and unsafe action by \(+0.0021\), with all observed-matrix intervals spanning zero. Model identity remains a role proxy, so the comparison is an explicit-role-feature ablation rather than complete role removal.

\section{Complete Contrast Ledger}
The six declared comparisons are single-teacher rationale KD minus SFT, unfiltered MTD minus single-teacher rationale KD, hard filter minus unfiltered MTD, correctness weighted minus hard filter, role-aware minus no-role weighted, and the primary correctness weighted minus unfiltered MTD. Each uses 10,000 crossed seed-by-group replicates with a fixed recorded random-number seed. Primary and component decision contrasts appear in Table~\ref{tab:2}. Other metrics remain diagnostic and are not promoted to grounding evidence.

\section{Human-Audit Protocol and Agreement}
The audit uses reference training seed 260631 and sampling seed 260701. Within each variant and stratum, a stable SHA-256 ordering of the variant, stratum, and source identifier is sampled without replacement. Gold labels and automatic scores do not determine eligibility or order. Positives remain 20 per arm. Safe controls use \(\min(20,\text{available})\); single-teacher rationale KD supplies 14 and each other arm supplies 20, for 160 positives and 154 safe rows.

The amendment was recorded after predictions were unsealed and the automatic summary had reached a fail-or-inconclusive state, but before human annotation. It changes only the safe-control availability rule. Safe rows contribute to the 314-row agreement calculation but not the positive-output counts in Table~\ref{tab:3}. Because each arm supplies positives from its own output policy and cross-arm source overlap was not serialized, the human results are non-paired descriptions.

Annotators use only the displayed task, evidence, allowed evidence identifiers, decision, answer, rationale, and citations. An answer is supported only when every material answer claim and the decision are licensed by the supplied evidence. A rationale is faithful only when every material proposition is evidence-supported and decision-consistent. Unsupported material is present when either answer or rationale contains a material claim not licensed by the evidence. A citation is valid only when every identifier is allowed and its evidence supports the associated material claim; an uncited positive material claim fails this field. Judgments are Yes, No, or Unclear, with Unclear conservatively treated as failure for support, faithfulness, and citation validity and as presence for unsupported material.

System variant, training seed, gold label, and automatic scores are hidden from the two independent annotators and the disagreement-only adjudicator. Nominal exact agreement and Cohen's \(\kappa\) use all 314 paired pre-adjudication rows. The arXiv source package includes the frozen instructions, amendment record, and machine-readable sampling configuration; it excludes identity mappings and restricted source text.

\section{Recalculation Scope and Integrity}
The internal recomputation uses hash-bound predictions, evaluator logic, group assignments, and fixed comparison configurations. It reproduces all 24 arm-seed point summaries, all six declared contrast jobs, and 906 comparable parsed result leaves within \(10^{-12}\). Of these, 889 parsed values are exactly equal and 17 differ only within floating-point tolerance; maximum absolute error is \(1.665\times10^{-16}\), with no leaf beyond tolerance.

Teacher generation, gate fitting, supervision construction, and student training are not rerun. Two historical Windows manifests are unavailable, so the audit validates the current observed inputs-to-report path rather than the complete upstream custody chain. The accompanying numerical-audit workbook records artifact hashes and a machine-readable evidence ledger without redistributing restricted source text.

\section{Metric Provenance Boundary}
Two legacy metric families are intentionally excluded from submission-level inference. A pooled F1 field is mechanically obtainable from current artifacts, but cross-version disagreements in its estimand, name, and qualification prevent its use in this manuscript. A risk-coverage quantity has unresolved provenance and estimand boundaries. Neither metric is reported numerically or used in the title, abstract, main tables, figures, discussion success claims, conclusion, or submission highlights.

\end{document}